\documentclass[10pt,letterpaper]{article}

\usepackage{cogsci}

\cogscifinalcopy 

\usepackage{pslatex}
\usepackage{apacite}
\usepackage{float} 

\usepackage{tabularx}
\usepackage{booktabs}
\usepackage{siunitx}
\usepackage{longtable}
\usepackage{multirow}
\usepackage{svg}
\usepackage{makecell}
\usepackage{caption}
\usepackage{amssymb}
\usepackage{subfig}
\usepackage{graphicx}
\usepackage{natbib}
\usepackage{svg}
\usepackage{mdframed}
\usepackage{listings}
\usepackage[symbol]{footmisc}

\title{How do Humans and Language Models Reason About Creativity? A Comparative Analysis}
\renewcommand{\thefootnote}{\fnsymbol{footnote}}
\author{{\large \bf Antonio Laverghetta Jr.\textsuperscript{1}\thanks{Corresponding author: $<$aml7990@psu.edu$>$} Tuhin Chakrabarty\textsuperscript{2} Tom Hope\textsuperscript{3} Jimmy Pronchick\textsuperscript{1} Krupa Bhawsar\textsuperscript{1} Roger E. Beaty\textsuperscript{1}} \\
  \textsuperscript{1}Pennsylvania State University \\
  \textsuperscript{2}Stony Brook University\\
  \textsuperscript{3}Hebrew University of Jerusalem \\
}

\begin{document}

\maketitle
\renewcommand*{\thefootnote}{\arabic{footnote}}
\setcounter{footnote}{0}

\begin{abstract}
Creativity assessment in science and engineering is increasingly based on both human and AI judgment, but the cognitive processes and biases behind these evaluations remain poorly understood. We conducted two experiments examining how including example solutions with ratings impact creativity evaluation, using a finegrained annotation protocol where raters were tasked with explaining their originality scores and rating for the facets of remoteness (whether the response is ``far'' from everyday ideas), uncommonness (whether the response is rare), and cleverness. In Study 1, we analyzed creativity ratings from 72 experts with formal science or engineering training, comparing those who received example solutions with ratings (example) to those who did not (no example). Computational text analysis revealed that, compared to experts with examples, no-example experts used more comparative language (e.g., ``better/worse'') and emphasized solution uncommonness, suggesting they may have relied more on memory retrieval for comparisons. In Study 2, parallel analyses with state-of-the-art LLMs revealed that models prioritized uncommonness and remoteness of ideas when rating originality, suggesting an evaluative process rooted around the semantic similarity of ideas. In the example condition, while LLM accuracy in predicting the true originality scores improved, the correlations of remoteness, uncommonness, and cleverness with originality also increased substantially --- to upwards of $0.99$ --- suggesting a homogenization in the LLMs evaluation of the individual facets. These findings highlight important implications for how humans and AI reason about creativity and suggest diverging preferences for what different populations prioritize when rating. 



\end{abstract}

\section{Introduction}
How do people evaluate and reason about creativity? In science and engineering, creativity assessment has traditionally relied on human experts to evaluate everything from grant proposals and scientific manuscripts to new technologies and engineering designs. Although experts routinely make these high-stakes decisions by weighing factors such as novelty and technical feasibility, the cognitive processes and biases that shape their evaluations are poorly understood. This challenge takes on new importance as artificial intelligence (AI) systems, particularly large language models (LLMs) \citep{brown2020language,ouyang2022training}, increasingly assume advanced roles in scientific research and innovation, from idea generation to peer review \citep{boiko2023emergent, d2024marg, lu2024ai,si2024can, wang-etal-2024-scimon}. LLMs can achieve impressive accuracy in predicting human creativity assessments \citep{organisciak2023beyond}, yet we know little about how they arrive at their judgments, what features they prioritize, or whether their evaluation strategies align with those of human experts. Understanding these cognitive and computational processes is crucial not only for advancing creativity research but also for developing AI systems that can better aid creative evaluation in STEM.

Modern creativity assessment is based on the Consensual Assessment Technique (CAT) \citep{amabile1982social,silvia2008assessing}, which uses human judgments from experts to reach a consensus on the creativity of a product/idea, often by assessing its originality or quality. In the standard implementation of this method, expert raters independently evaluate creative products without specific scoring criteria, relying on their implicit domain understanding. While this approach has demonstrated remarkable reliability and predictive validity --- CAT ratings predict real-world creative achievements across multiple domains \citep{silvia2008assessing} --- it typically relies on global originality scores that may mask underlying evaluation processes \citep{Cseh2019}. Notably, originality itself can be understood as an aggregation of three distinct facets: uncommonness, remoteness, and cleverness \citep{silvia2008assessing}, though their relative contributions to expert judgments remain unclear. Understanding these evaluation patterns is especially crucial for STEM domains, where creative solutions must carefully balance novelty with real-world technical constraints. Even studies of simpler ideation tasks like the Alternate Uses Task (AUT) suggest this complexity, showing how different factors like novelty and appropriateness contribute distinctly to creativity judgments \citep{Diedrich2015}.


The cognitive processes underlying creativity evaluation are becoming clearer through recent empirical work \citet{wang2025behind}. Early think-aloud studies like \citet{gilhooly2007divergent} focused on idea-generation processes in the AUT, showing how participants move from memory retrieval to more abstract strategies. More recently, \citet{orwig2024creative} used linguistic inquiry and word count (LIWC) analysis, a computerized text analysis method that quantifies psychological dimensions of language \citep{tausczik2010psychological}, to analyze how participants explain their originality ratings on the AUT. Results revealed that even when judges agree on creativity scores, they often employ different cognitive processes in their evaluations, as evidenced by variations in their use of memory-related terms, temporal focus (past vs. future orientation), and analytical language.

STEM creativity represents a crucial yet understudied domain for creativity assessment. Although extensive research has examined scientific hypothesis generation and engineering design thinking, we know surprisingly little about how experts evaluate creative merit in STEM contexts, where ideas must balance novelty with technical feasibility. To assess creative thinking in STEM, \citet{Patterson2025} developed a novel Design Problems Task (DPT) that measures the ability to generate solutions to real-world STEM challenges, capturing the dual constraints that characterize expert-level scientific and engineering creativity without the need for expert level knowledge to solve the items. The DPT spans three domains: ability difference and limitations (e.g., assisting people with learning impairments), transportation and mobility (e.g., reducing traffic congestion), and social environments and systems (e.g., improving access to clean water), with participants generating multiple solutions that are then rated for both originality and effectiveness.

The demanding nature of creativity evaluation, requiring expert raters to assess thousands of open-ended responses while maintaining consistent judgment standards, has sparked interest in automating assessment using AI. Recent work has demonstrated that LLMs trained on human creativity ratings can achieve impressive accuracy in evaluating novel responses \citep{organisciak2023beyond,Patterson2025}. Like human experts, these models may be influenced by contextual information and examples provided during evaluation. However, these AI evaluators present their own interpretability challenges. While AI may agree with human ratings, we know little about how they arrive at their judgments, what features of responses they attend to during evaluation, or whether those features are similar to those attended to by human experts. Understanding AI creativity evaluation mechanisms for STEM-related tasks has become increasingly urgent as LLMs take on expanded roles in scientific research, from peer review \citep{huang2025large,lin2024evaluating} to idea generation \citep{gu2024interesting,si2024can}. If these models are to serve as reliable creativity judges, we must better understand their evaluation processes and potential biases. This knowledge could inform both model interpretability efforts and attempts to align AI creativity assessment with human judgment. Moreover, improving evaluation capabilities may enhance idea generation abilities, consistent with cognitive models that link these processes \citep{smith1995creative}. 


The present research examines how human experts and LLMs evaluate STEM creativity through real-world design problems. In two studies, we conduct a fine-grained analysis of the factors influencing creativity assessment in STEM domains. First, we examine how human experts evaluate originality in STEM solutions, analyzing both their numerical ratings and their written explanations to understand the cognitive processes involved. We also investigate how providing examples impacts expert judgment, testing whether contextual information (example design solutions and originality ratings) changes how experts weigh different aspects of creativity. Second, we perform a parallel analysis of LLMs, examining whether these models show similar patterns in their creativity evaluations and exploring potential differences between human and AI assessment strategies.\footnote{Code, data, and supplementary materials are available at: https://github.com/Beaty-Lab/CogSci-2025-Scientific-Creativity}

\section{Study 1: Human Creativity Evaluations}
Our first study sought to understand the key factors underlying human expert evaluation of the creativity of solutions to design problems (DPT) items. A participant in this task is given a scientific or engineering problem (e.g., increasing the use of renewable energy) and is instructed to come up with as many novel solutions to the problem as they can think of. Similar to expert-level science, the best solutions are both original and feasible, though unlike other STEM assessments the DPT benefits from but is not contingent on expertise to come up with creative ideas. The greater complexity of DPT responses compared to those from other creativity tests and its relationship to scientific creativity more broadly makes it a strong choice for our analysis. Unlike prior studies, which often have experts rate only the originality or quality of products, we instead ask our raters to provide fine-grained assessments of cleverness (whether the solution is insightful or witty), remoteness (whether the solution is ``far'' from everyday ideas), and uncommonness (whether the solution is rare, given by few people) in addition to originality, each of which is thought to influence ratings of creativity \citep{silvia2008assessing}. These assessments are performed both with and without the presence of example creativity ratings to DPT items, enabling us to examine how added context affects the evaluation process. Finally, we ask experts to briefly explain their originality scores, enabling us to employ methods from computational text analysis to probe the cognitive processes experts employ when rating and how such processes may be modulated by added context.

\subsection{Methods}
We use the data from \citet{Patterson2025}, who obtained more than 7000 responses to DPT items from undergraduate STEM majors. Each response was rated for originality using a five-point Likert scale by at least three expert raters with formal training in engineering. We drop items that did not obtain at least one rating from every point of the scale (certain items never had a response that received a five). We convert Likert scores into factor scores, as this has been shown to provide more accurate creativity ratings \citep{silvia2008another}, and we treat these factor scores as the ground truth originality scores of each response.

We recruit 80 participants on Prolific to provide finegrained creativity ratings to DPT responses, requiring that they have a bachelor's degree or higher in a STEM field and are fluent in English. We split participants into two conditions: a \textit{no example} condition where participants are given responses to rate without any additional context, and an \textit{example} condition where participants are first shown example solutions with originality scores for responses to the same prompt being rated. We pull three example solutions from the same dataset while ensuring that participants never rate them. We include a solution with a score of one, one with a score of three, and one with a score of five, to avoid biasing participants towards either end of the scale. We first have each participant rate for originality following the same procedure, instructions, and facet definitions as \citet{Patterson2025}. After rating originality, participants in both groups then provide 1-2 sentences explaining their rating process \citep{orwig2024creative}, and they finish by rating the uncommonness, remoteness, and cleverness of the response using a five-point Likert scale for each. We instruct participants to be specific in their explanations, to draw on their domain expertise as holders of a STEM degree, and to avoid overly simplistic explanations (e.g., ``it's not original'' or ``it's an obvious answer''). We define a good explanation as being at least one sentence long and including specific details from the participant's prior experience, the response, or the examples (if applicable). We also provide definitions of uncommonness, remoteness, and cleverness for the final rating task, emphasizing that each facet is related while being distinct from originality. We include educational background and AI use checks at the end of the survey.

We administer each participant 15 DPT responses at random. To encourage high-quality explanations, we offer \$20 per hour to complete a 30-minute study. We exclude participants with an approval rating of less than 90\%, who report using AI to complete the task, or who report an education level lower than the minimum specified on Prolific. We also exclude participants who were exceptionally slow or fast (with a completion time further than three standard deviations from the mean), who gave the same rating for every response, or who did not follow our instructions for formatting explanations (as checked by a research assistant). This resulted in a final sample size of 37 participants and 481 ratings in the example condition and 35 participants and 455 responses for the no example.


When examining the participants' explanations, we employ an analysis plan similar to \citet{orwig2024creative}, who used LIWC to analyze explanations of originality scores for AUTs. However, recent work has found that LLMs can predict psycholinguistic features of text more strongly than LIWC, even zero shot \citep{rathje2024gpt}. Therefore, we use LLMs to automatically rate linguistic markers in the explanations. We instruct LLMs to rate for the following variables: 

\begin{itemize}
    \item \textit{Past/future expressions}: Is the explanation past-focused or future-focused in its evaluation of the response?
    \item \textit{Perceptual details}: Does the explanation focus on the process of perceiving (``observe'', ``seen'', ``heard'', ``feel'', etc.)?
    \item \textit{Causal/analytical}: Does the explanation involve a structured evaluation of the response, evidencing an analytical process, or is the explanation more intuitive in its justifications?
    \item \textit{Comparative}: Does the explanation make explicit references to standards or examples or compare the response to other ideas?
    \item \textit{Cleverness}: Does the explanation refer to the cleverness, wittiness, shrewdness, or ingenuity (or lack thereof) of the response?
\end{itemize}

Both past/future language use and perceptual details have been explored to assess cognitive strategies employed on other creativity tests \citep{orwig2024creative}. We elect to use causal/analytical, comparative, and cleverness linguistic markers to aid in assessing whether participants employed a more structured process --- which might be evidenced by causal/analytical or comparative language use --- or a more intuitive process, as evidenced by language indicating sensory experiences or other ``gut reactions'' (e.g, ``it feels like a clever idea''). These linguistic markers also map onto the finegrained facets participants were asked to rate, with cleverness language mapping onto cleverness and comparative language mapping onto remoteness and uncommonness (as both remoteness and uncommonness often require making references to prior solutions). We use both \textsc{claude-3.5-sonnet}\footnote{https://www.anthropic.com/news/claude-3-5-sonnet} and \textsc{gpt-4o}\footnote{https://openai.com/index/hello-gpt-4o/} to check for reliability in ratings and avoid biases specific to a single LLM, though due to space constraints we mainly report results from \textsc{gpt-4o} as this is the model \citet{rathje2024gpt} validated. To encourage deterministic output, we set the temperature for both models to $0$ and top P to $1$. We instruct LLMs to rate each facet and provide a binary evaluation of whether the explanation does or does not contain the feature. Prompts are provided in the supplementary materials.

\subsection{Results}
\begin{figure}[htb]
    \centering
    \footnotesize
    \includegraphics[width=0.8\linewidth]{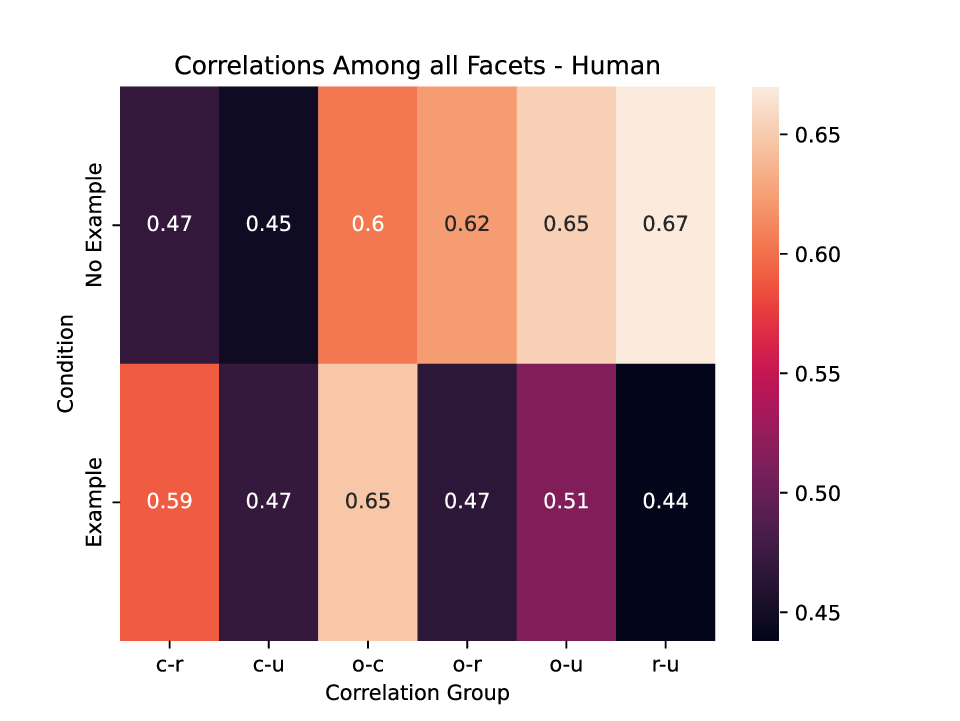}
    \caption{Pearson correlations among pairwise Likert ratings for both conditions. o = originality, c = cleverness, u = uncommonness, r = remoteness.}
    \label{fig:experiment_1_correlations}
\end{figure}

We begin by examining inter-correlations among all facets (cleverness, remoteness, uncommonness) and correlations between each facet and originality for both conditions. Results are in Figure \ref{fig:experiment_1_correlations}. As expected, each facet is moderately correlated with originality as well as each other, with Pearson r in the range 0.45--0.67 (all correlations are significant).\footnote{Results from all correlational analysis in both studies were similar using Spearman $\rho$.} Comparing the example to no example conditions, we see an increase in correlation between originality and cleverness and a decrease in correlation between originality and both remoteness and uncommonness. Changes in correlation across conditions were significant for cleverness-remoteness (Fisher's z = 2.83, p $<$ 0.01), remoteness-uncommonness (z = -4.61, p $<$ 0.001), and remoteness-originality (z = -2.96, p $<$ 0.01), but were insignificant for all other comparisons. Notably, the presence of the examples did not make experts significantly more accurate in their evaluations of originality, with correlations in the moderate range for both conditions (no example r = 0.44, example r = 0.47).




Turning to participant explanations, \textsc{gpt-4o}'s ratings did not reveal significant differences per condition for perceptual details, past/future language use, or cleverness, but differences are significant for both causal/analytical language (Mann-Whitney U = 78039.5, p $<$ 0.05) and comparative language (U = 75627.5, p $<$ 0.01) with the example condition using less comparative and causal/analytical language than the no examples. Distributions for linguistic markers are shown in Figure \ref{fig:liwc_analysis}. \textsc{claude-3.5-sonnet}'s ratings generally agreed with \textsc{gpt-4o} (Cramer's V in the range 0.549--0.798) with the only notable departure being that \textsc{claude-3.5-sonnet} found no significant difference in causal/analytical language between the conditions (U = 75076.5, p $<$ 0.5). We report additional linguistic marker analysis in the supplementary materials.


\subsection{Discussion}
As expected, the facet ratings did not correlate above 0.67 for any pair, implying that participants at least partially distinguished among each facet when assessing originality. Further, correlations changed by a significant degree when including example ratings, with both remoteness and uncommonness becoming weaker predictors of originality and cleverness becoming a stronger one. Given that participants in the no example condition needed to actively retrieve example solutions from memory when evaluating, a possible explanation is that this retrieval process biased them towards placing stronger emphasis on the remoteness and uncommonness of the response in relation to solutions they had seen in the past, while example participants would instead have these cognitive resources free for other aspects of evaluation. Notably, participants in both groups did not differ significantly in terms of education, making it unlikely this effect could be explained as a skill confound. The idea that participants in the example condition were biased toward cleverness rather than the other facets was also partially supported by their explanations, as no example participants used significantly more comparative language than example participants. Given that assessing remoteness or uncommonness often requires making direct comparisons to prior solutions, it makes sense that an evaluation rooted around these facets would contain more comparisons than an evaluation rooted around cleverness, which is more readily evaluated in isolation (e.g., whether the idea is resource efficient, not immediately obvious, etc.).

\section{Study 2: LLM Creativity Evaluations}
Our analysis thus far has shed light on how experts reason about originality when taking on the role of creativity evaluators and how this is affected by the inclusion of context. But do these same effects hold for LLMs when they are used to evaluate creativity? Conceptually, the examples given to humans serve a similar role as few-shot learning, a well-established method of improving the accuracy of LLMs on classification tasks \citep{brown2020language}. Yet exemplars may also bias models in other ways, especially in technically complex domains like science and engineering where the nature of truly creative ideas can be complex and difficult to discern a priori \citep{schmidt2011creativity,simonton2004creativity}. Scientific evaluations given by LLMs also tend to be markedly different from humans, with LLMs often overestimating the quality of scientific research \citep{schmidgall2025agent}. Given the increasing role generative AI is having both in expert-level science and in STEM education, it is crucial to perform a similar finegrained analysis of LLM creativity evaluations to enable a head-to-head comparison between human experts and LLMs as evaluators. Our second experiment sought to perform this comparison, using the same methods as in the first experiment but using ratings from LLMs.


\subsection{Methods}
We use \textsc{claude-3.5-haiku} and \textsc{gpt-4o-mini}, as these LLMs tend to achieve competitive performance on AI benchmarks while also being cost-effective \citep{chiang2024chatbot}. Further, because we use the larger variants of OpenAI and Anthropic models to rate explanations (\textsc{claude-3.5-sonnet} and \textsc{gpt-4o}), we chose to use these smaller variants for this study to avoid possible biases from LLMs recognizing their own output \citep{panickssery2024llm}. We set the temperature to $0$ and top P to $1$ for all trials while leaving other hyperparameters at their defaults. We structure our prompt similar to the instructions given to the human participants. We instruct the LLM to rate originality, cleverness, remoteness, and uncommonness and to explain its originality score. LLMs are given no exemplars in the no example condition and are given the same examples as humans in the example condition. We administer the same datasets for both conditions as we used in the first experiment, including duplicate archival responses, to make results from humans and LLMs as comparable as possible.\footnote{Note that, even with temperature set to zero, these LLMs may generate different ratings for the same response.} We include our prompts in the supplementary materials.

\subsection{Results}
\begin{figure}[htb]
    \centering
    \footnotesize
    \includegraphics[width=0.8\linewidth]{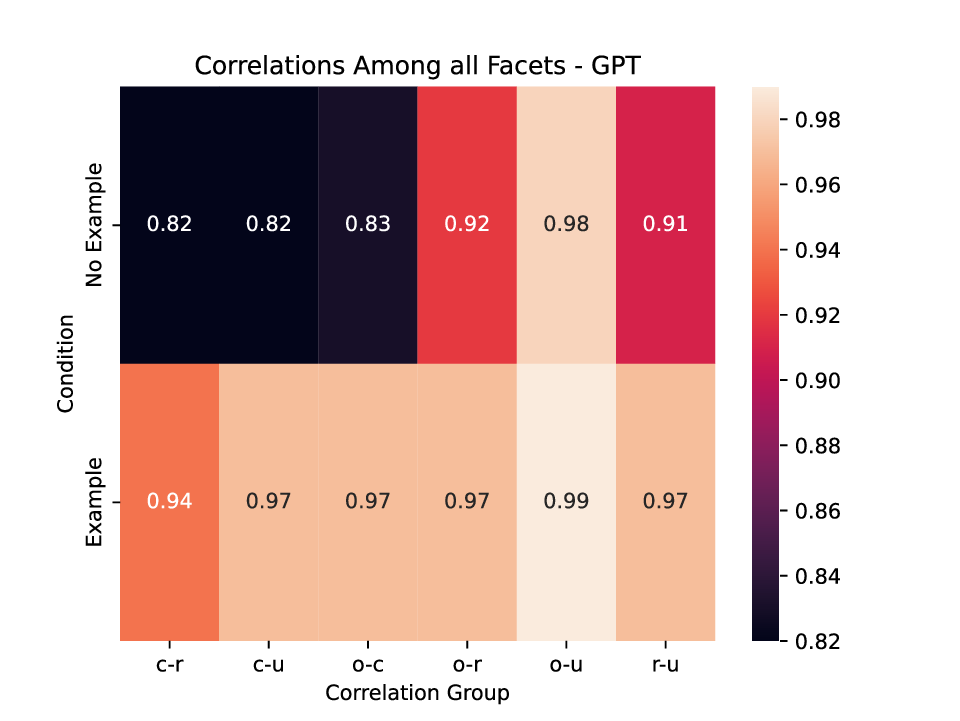}
    \caption{Pearson correlations among pairwise Likert ratings for \textsc{gpt-4o-mini} in both conditions. o = originality, c = cleverness, u = uncommonness, r = remoteness.}
    \label{fig:experiment_2_gpt_correlations}
\end{figure}

\begin{figure}[htb]
    \centering
    \footnotesize
    \includegraphics[width=0.8\linewidth]{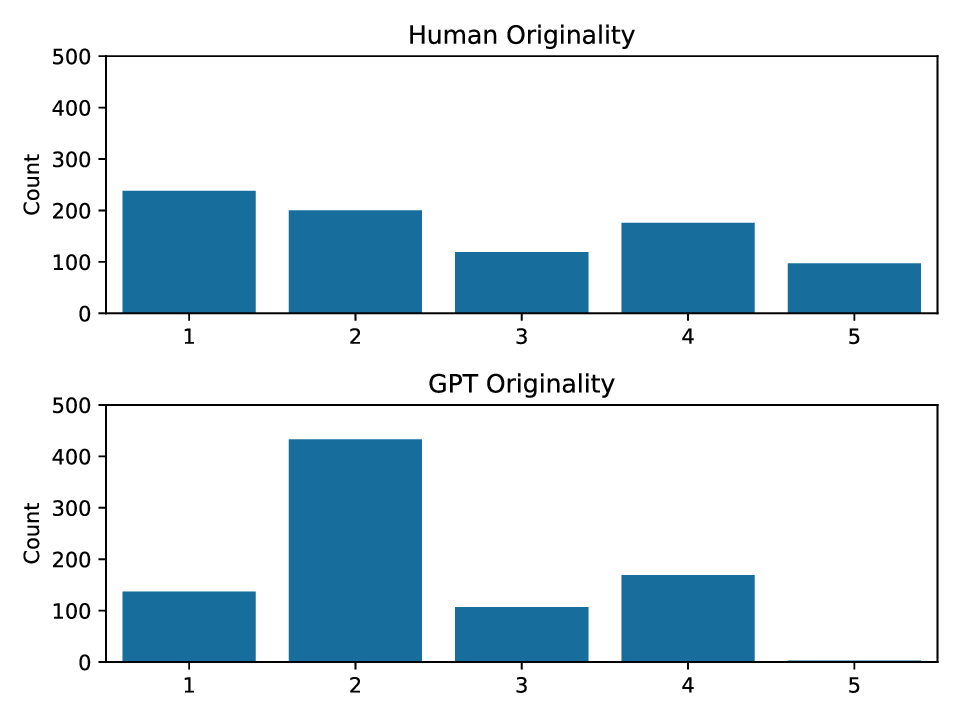}
    \caption{Human and \textsc{gpt-4o-mini} originality scores.}
    \label{fig:originality_distros}
\end{figure}

LLM originality predictions correlated strongly with the ground truth, and examples significantly boosted this correlation (no example: r = $0.6$, $0.67$; example: r = $0.74$, $0.76$; all correlations significant). \textsc{claude-3.5-haiku} and \textsc{gpt-4o-mini} exhibited strong agreement in their ratings, with correlations between their facet scores in the range $0.73$--$0.88$. Figure \ref{fig:experiment_2_gpt_correlations} summarizes facet correlations for \textsc{gpt-4o-mini}. Cleverness was the weakest predictor of originality scores in the no example condition, with remoteness and uncommonness being much more strongly correlated with originality. However, this effect dissipated in the example condition, with the strength in correlation between originality and each facet increasing significantly (all changes in correlations were found significant using Fisher's z test). To account for the possibility of these correlations being influenced by chance agreement among the facets, we also performed the same comparisons using Cohen's Kappa and obtained similar findings, detailed results are in the supplementary materials. We compare the distributions of human and \textsc{gpt-4o-mini} originality scores across both conditions in Figure \ref{fig:originality_distros}. We find that LLMs rarely use a five and rate a majority of responses as a two. 

\begin{figure*}[t]
    \centering
    \footnotesize
    \subfloat{
        \includegraphics[width=0.8\textwidth]{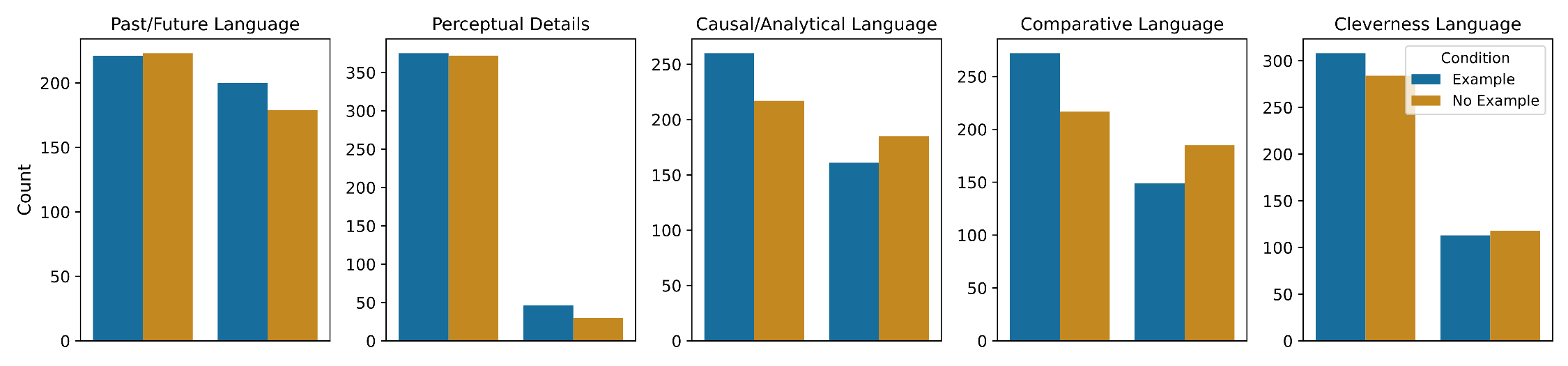}
        \label{fig:human_cleverness}
    }
    \hspace{-2cm}
    \subfloat{
        \includegraphics[width=0.8\textwidth]{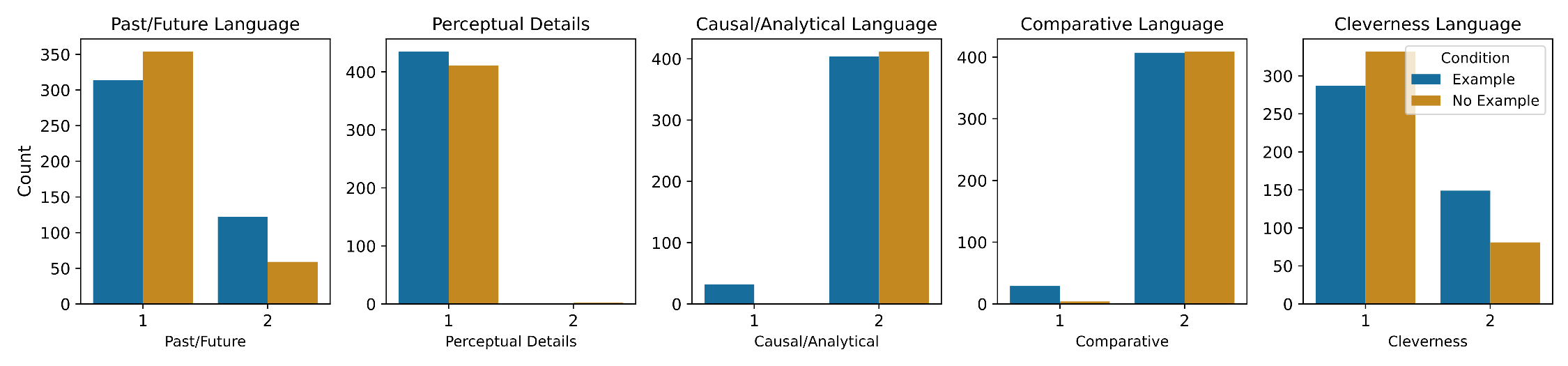}
        \label{fig:gpt_cleverenss}
    }
    \caption{Comparison between linguistic marker use from humans (top) and \textsc{gpt-4o-mini} (bottom), as assessed by \textsc{gpt-4o}. A rating of 1 indicates the feature is absent in the response, 2 indicates it is present.}
    \label{fig:liwc_analysis}
\end{figure*}

Figure \ref{fig:liwc_analysis} compares human and \textsc{gpt-4o-mini} explanations. For \textsc{gpt-4o-mini}, both \textsc{gpt-4o} and \textsc{claude-3.5-sonnet} found no significant difference in perceptual details per condition, but did find significant differences in use of cleverness language (Claude U = 106495, p $<$ 0.001; GPT U = 1031445.5, p $<$ 0.001), with models producing conflicting ratings for the remaining markers (results were similar for \textsc{claude-3.5-haiku}). Comparing LLMs to humans, we find more variability in the presence or absence of all linguistic markers in humans as opposed to LLMs, with LLMs having much more heavily skewed rating distributions and sometimes having a marker completely absent across all explanations, which never occurred in humans. LLM explanations also tended to follow a more rigidly analytical structure, implying a more structured evaluation compared to humans, who had more instances of perceptual details that implied a more intuitive process drawing on memory. Results for \textsc{claude-3.5-haiku} are similar and are included in the supplementary materials.


\subsection{Discussion}
Our findings highlight key differences in how LLMs reason about creativity as compared to humans. While human judges appeared to more strongly associate originality with cleverness, the opposite pattern emerged in LLMs, where originality was more strongly associated with remoteness and uncommonness. Though one might be inclined to trust the LLM originality evaluations more, given their stronger correlation with the true ratings, this should be balanced against the LLMs' apparent inability to distinguish among cleverness, remoteness, or uncommonness while rating. Human facet correlations were consistently much weaker than they were for LLMs, which reflects the conceptualization of these facets as impacting originality while not being the same construct. This lies in stark contrast to the LLMs' much stronger facet correlations --- in some cases nearing perfect correlation with originality in the example condition. It is especially noteworthy that differences in facet correlations appeared washed out by the examples, even as model predictions became more accurate. While LLM originality scores had stronger predictive validity, they also had weaker construct validity due to a homogenization of cleverness, remoteness, and uncommonness, which unlike for humans was only strengthened by the presence of the examples. 

Our analysis of LLM explanations pointed to similar discrepancies between humans and AI. LLMs exhibited less diversity in explanation styles, with much more heavily skewed distributions of linguistic markers as compared to humans. This mirrors both the distributions of Likert originality scores, and similar findings in other areas of creativity and the social sciences, where generative AI has been found to suffer from less diversity in generations \citep{park2024diminished}. This trend may have partially been driven by our use of a low temperature value, though we note that exact duplicate explanations were uncommon even for the same problem (especially for \textsc{gpt-4o-mini}), making it unlikely this is the sole reason for redundancy. LLMs did align with humans along several linguistic markers, with both tending to not employ perceptual details and using less future-focused language in the no example condition as compared to the example condition, though not all such trends were statistically significant and LLMs did not closely align with humans for the majority of the linguistic markers explored. It appears that, like with humans, the examples qualitatively impacted LLM explanations, leading to shifts in the content of the explanations as opposed to the no example, which is an important consideration when LLMs are used to evaluate creativity in the real world.

\section{Conclusion}
Understanding how AI reasons about the creativity of products and whether that reasoning process aligns with human experts becomes ever more paramount as AI assumes the role of creativity evaluators. Although many works have studied LLMs' ability to accurately rate originality across various tasks \citep{lin2024evaluating,schmidgall2025agent}, the coarse-grained nature of such evaluations makes drawing conclusions about their rating process difficult. We contribute to this literature by collecting finegrained originality evaluations of responses to science and engineering prompts from both human experts and LLMs. Our analysis reveals substantial differences in how these populations both rate for originality in relation to other facets and in how they structure explanations of their rating process. It appears that how LLMs are affected by context and how they rate individual facets of originality is markedly different from human judges, carrying important implications for deploying LLMs as evaluators. Notably, these differences point to potential validity issues with the LLM scores, with AI producing vastly different facet correlations despite more accurately predicting ground truth originality scores. While it is possible that such differences may fade as new LLMs are developed, identifying and understanding these discrepancies remains crucial for advancing work in automated creativity scoring, given that current LLMs already excel at creativity scoring at a coarse-grained level \citep{organisciak2023beyond,luchini2025automated}.


Future work can expand on our contributions in several ways. Due to budget constraints, we were unable to run multiple pairwise comparisons between humans and LLMs under multiple instruction sets and prompt variations, which is an important step for quantifying the sensitivity of LLM ratings to the input prompt. Though we mitigated this by using multiple LLM evaluators, it remains possible that our results may have been driven in part by the structure of the prompt. Similarly, due to time constraints in human studies, we selected only a small number of examples and were unable to run studies that varied either the examples themselves or their order of presentation. It is possible that this could have biased LLM judgments of cleverness, uncommonness, and remoteness, given that we were unable to include multiple examples at each originality rating. Though we do not believe this could fully explain the homogenization effect, since correlations among the facets were already stronger than in humans zero shot, it remains an important analysis to help us further understand this effect.

Creativity is often considered one of the most critical skills to master in modern economies \citep{illessy2020automation,tsegaye2019antecedent}, and AI must be developed to evaluate it both accurately and fairly. Achieving this goal requires understanding how AI arrives at creativity judgments and whether it rates individual facets of creativity in the same way as humans. We hope our work provides insights into how AI reasons about creativity and serves as a call to action to perform similar finegrained assessments of creativity in other domains.

\section{Acknowledgments}
R.E.B. is supported by grants from the National Science Foundation [DRL-1920653; DRL-240078; DUE-2155070].

\bibliographystyle{apacite}

\setlength{\bibleftmargin}{.125in}
\setlength{\bibindent}{-\bibleftmargin}

\bibliography{antonio}

\clearpage
\appendix
\section{LLM Prompts}
We structure the prompts for LLM-based analysis of explanations similar to \cite{rathje2024gpt}, who gave models minimal context when rating LIWC categories. Prompts for past/future language, perceptual details, causal/analytical language, comparative language, and cleverness are shown in Figures \ref{fig:past-future-prompt}, \ref{fig:perceptual-details-prompt}, \ref{fig:causal-analytical-prompt}, \ref{fig:comparative-prompt}, and \ref{fig:cleverness-prompt} respectively. Our prompt for Study 2 is listed in Figure \ref{fig:experiment-2-prompt}.

\begin{figure*}[htbp]
    \begin{mdframed}[linewidth=0.5pt]
        \begin{lstlisting}[
            basicstyle=\ttfamily\small,
            breaklines=true,
            frame=none
        ]
Are the words in this text past-focused or future-focused? ANSWER ONLY WITH A NUMBER: 1 if past-focused, 2 if future-focused. Here is the text: [RESPONSE]
        \end{lstlisting}
    \end{mdframed}
    \caption{Past/future language prompt. [RESPONSE] is filled with the participant explanation.}
    \label{fig:past-future-prompt}
\end{figure*}

\begin{figure*}[htbp]
    \begin{mdframed}[linewidth=0.5pt]
        \begin{lstlisting}[
            basicstyle=\ttfamily\small,
            breaklines=true,
            frame=none
        ]
Perceptual details refer to the process of perceiving and are indicated by words like observing, seen, heard, feeling, listen, or touch, among others. Does this text contain perceptual details? ANSWER ONLY WITH A NUMBER: 1 if the text doesn't contain perceptual details, 2 if the text does contain perceptual details. Here is the text: [RESPONSE]
        \end{lstlisting}
    \end{mdframed}
    \caption{Perceptual details prompt. [RESPONSE] is filled with the participant explanation.}
    \label{fig:perceptual-details-prompt}
\end{figure*}

\begin{figure*}[htbp]
    \begin{mdframed}[linewidth=0.5pt]
        \begin{lstlisting}[
            basicstyle=\ttfamily\small,
            breaklines=true,
            frame=none
        ]
Casual/analytical markers refer to performing structured evaluation of ideas. The cause/insight/analytical LIWC category words are example markers of causal processes. Causual markers are contrasted with intuitive markers, which do not indicate a structured evaluation and instead mark an intuitive (e.g., gut feeling) one. Does this text contain causal/analytical markers? ANSWER ONLY WITH A NUMBER: 1 if the text uses an intuitive and not analytical approach, 2 if the text uses a structured and analytical approach. Here is the text: [RESPONSE]
        \end{lstlisting}
    \end{mdframed}
    \caption{Causal/analytical prompt. [RESPONSE] is filled with the participant explanation.}
    \label{fig:causal-analytical-prompt}
\end{figure*}

\begin{figure*}[htbp]
    \begin{mdframed}[linewidth=0.5pt]
        \begin{lstlisting}[
            basicstyle=\ttfamily\small,
            breaklines=true,
            frame=none
        ]
Comparative language tracks explicit references to standards or examples and includes comparing or contrasting ideas or drawing distinctions among ideas. Does this text contain comparative language? ANSWER ONLY WITH A NUMBER: 1 if the text doesn't contain comparative language, 2 if the text does contain comparative language. Here is the text: [RESPONSE]
        \end{lstlisting}
    \end{mdframed}
    \caption{Comparative language prompt. [RESPONSE] is filled with the participant explanation.}
    \label{fig:comparative-prompt}
\end{figure*}

\begin{figure*}[htbp]
    \begin{mdframed}[linewidth=0.5pt]
        \begin{lstlisting}[
            basicstyle=\ttfamily\small,
            breaklines=true,
            frame=none
        ]
Does this text refer to the cleverness, wittiness, shrewdness, or ingenuity of an idea? ANSWER ONLY WITH A NUMBER: 1 if the text doesn't refer to cleverness, 2 if the text does refer to cleverness. Here is the text: [RESPONSE]
        \end{lstlisting}
    \end{mdframed}
    \caption{Cleverness prompt. [RESPONSE] is filled with the participant explanation.}
    \label{fig:cleverness-prompt}
\end{figure*}

\begin{figure*}[htbp]
    \begin{mdframed}[linewidth=0.5pt]
        \begin{lstlisting}[
            basicstyle=\ttfamily\small,
            breaklines=true,
            frame=none
        ]
You are evaluating the originality of solutions to this engineering design problem: [PROBLEM]

You will rate solutions on multiple criteria using 1-5 scales and provide explanations. Your response must follow this exact format:

ORIGINALITY: [1-5 or NA]
UNCOMMON: [1-5]
REMOTE: [1-5]
CLEVER: [1-5]
EXPLANATION: [1-2 sentences explaining the originality rating from a STEM perspective]

Rating scales:
Originality:
1: Very Unoriginal
2: Unoriginal
3: Neutral
4: Original
5: Very Original
NA: For solutions that are too short or unclear

Uncommon:
1: Very common
2: Common
3: Neutral
4: Uncommon
5: Very uncommon

Remote:
1: Very unremote
2: Unremote
3: Neutral
4: Remote
5: Very remote

Clever:
1: Very unclever
2: Unclever
3: Neutral
4: Clever
5: Very clever

Evaluation criteria:
- Uncommon: Consider if the solution is rare. Common solutions given by many people score low. A unique solution may score high, unless it's just strange rather than original.
- Remote: Consider how far the solution is from everyday ideas. Non-obvious solutions score higher. Solutions similar to common approaches score lower.
- Clever: Consider the insight and wit shown. Clear, insightful solutions score higher. Unclear solutions score lower. Even common solutions can score high if presented cleverly.

Provide your STEM-based explanation focusing solely on the originality rating. Be specific about why the solution is or isn't original from an engineering perspective.
        \end{lstlisting}
    \end{mdframed}
    \caption{Prompt for LLM data collection in Study 2. [PROBLEM] is filled with the design problem.}
    \label{fig:experiment-2-prompt}
\end{figure*}

\section{Additional Correlation and Distribution Results}
\begin{figure}[H]
    \centering
    \footnotesize
    \includegraphics[width=1\linewidth]{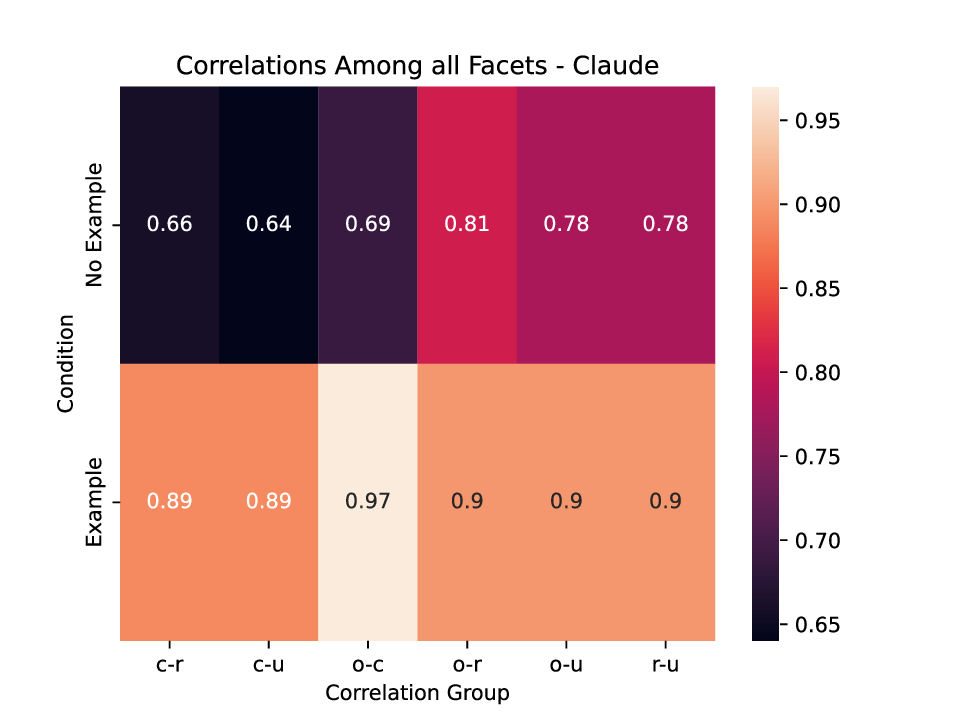}
    \caption{Pearson correlations among pairwise Likert ratings for \textsc{claude-3.5-haiku} in both conditions. o = originality, c = cleverness, u = uncommonness, r = remoteness.}
    \label{fig:experiment_2_claude_correlations}
\end{figure}

\begin{figure}[H]
    \centering
    \footnotesize
    \includegraphics[width=1\linewidth]{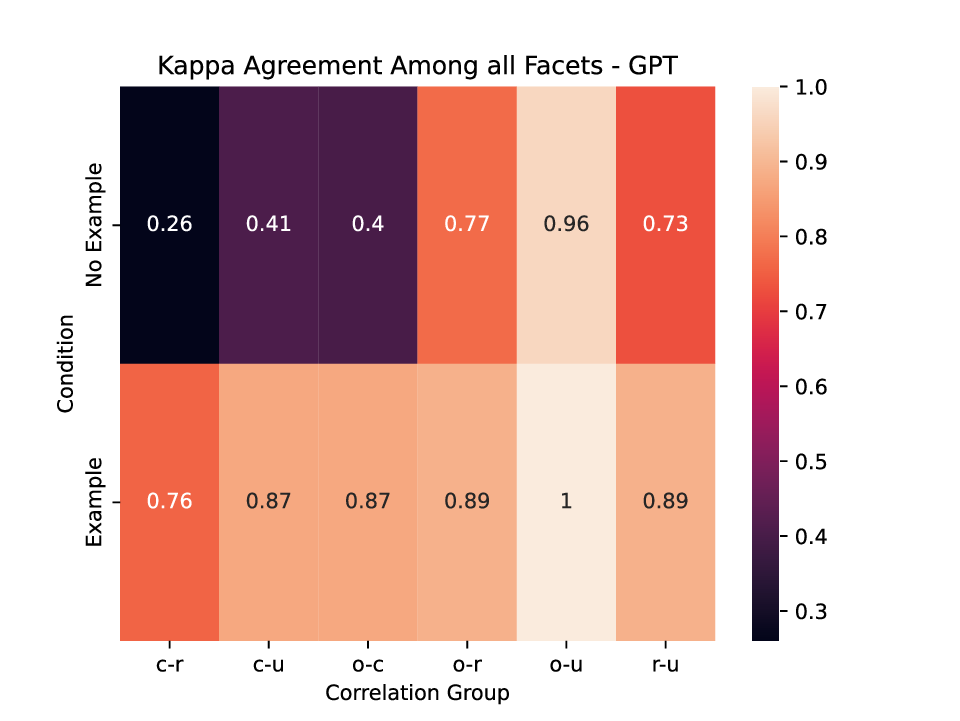}
    \caption{Kappa agreement statistics \textsc{gpt-4o-mini} in both conditions. o = originality, c = cleverness, u = uncommonness, r = remoteness.}
    \label{fig:gpt_agreement}
\end{figure}

\begin{figure}[H]
    \centering
    \footnotesize
    \includegraphics[width=1\linewidth]{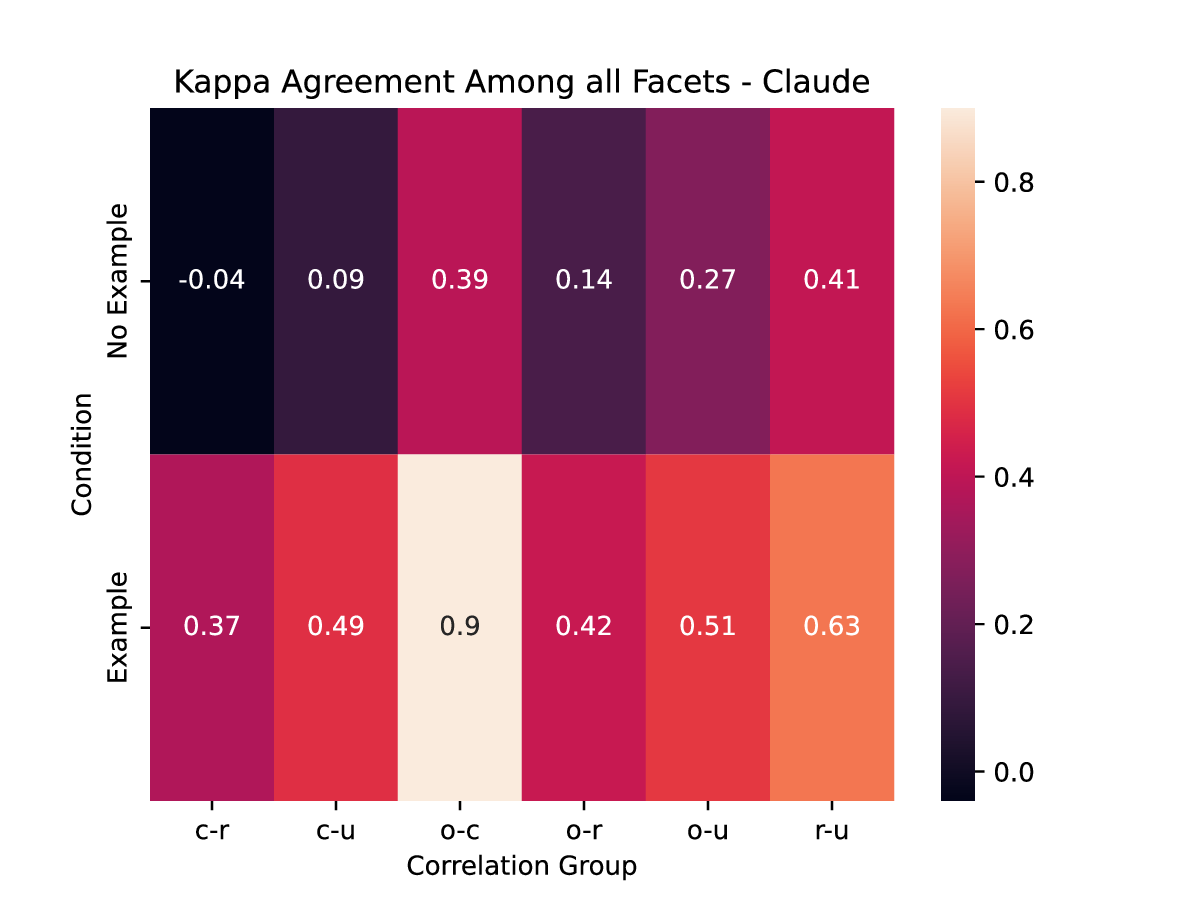}
    \caption{Kappa agreement statistics \textsc{claude-3.5-haiku} in both conditions. o = originality, c = cleverness, u = uncommonness, r = remoteness.}
    \label{fig:claude_agreement}
\end{figure}

\begin{figure*}[t]
    \centering
    \footnotesize
    \subfloat{
        \includegraphics[width=0.5\textwidth]{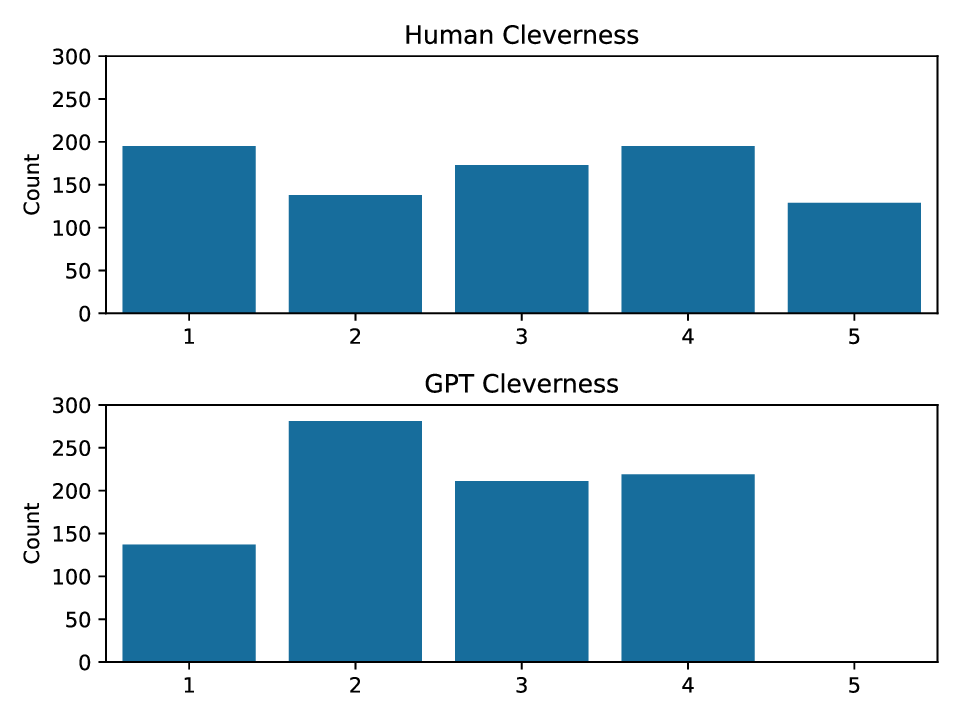}
    }
    \hfill
    \subfloat{
        \includegraphics[width=0.5\textwidth]{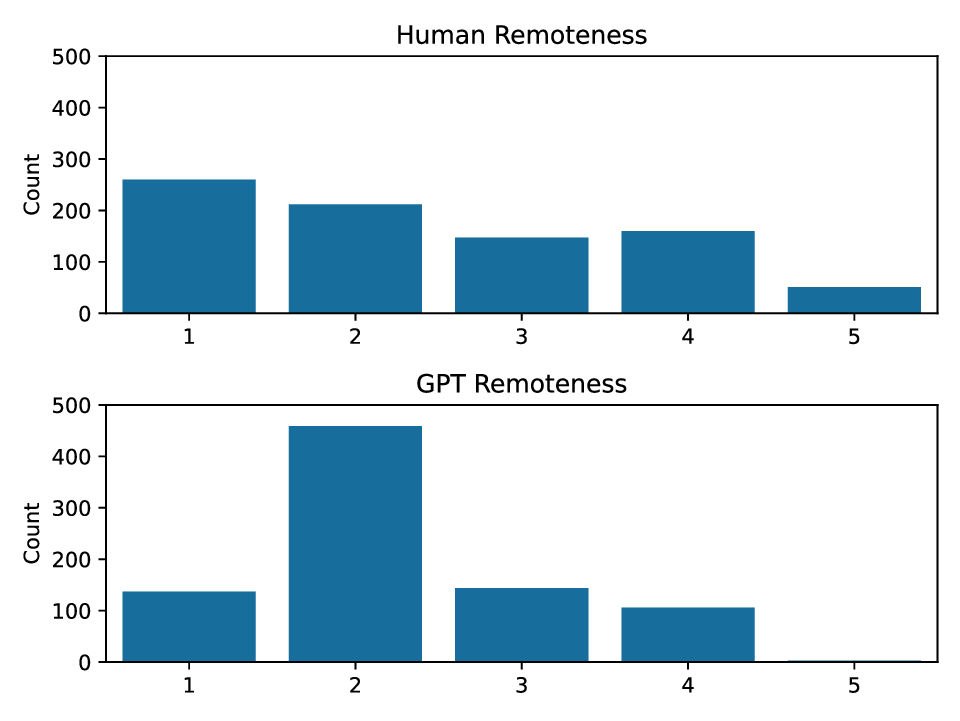}
    }
    \hfill
    \subfloat{
        \includegraphics[width=0.5\textwidth]{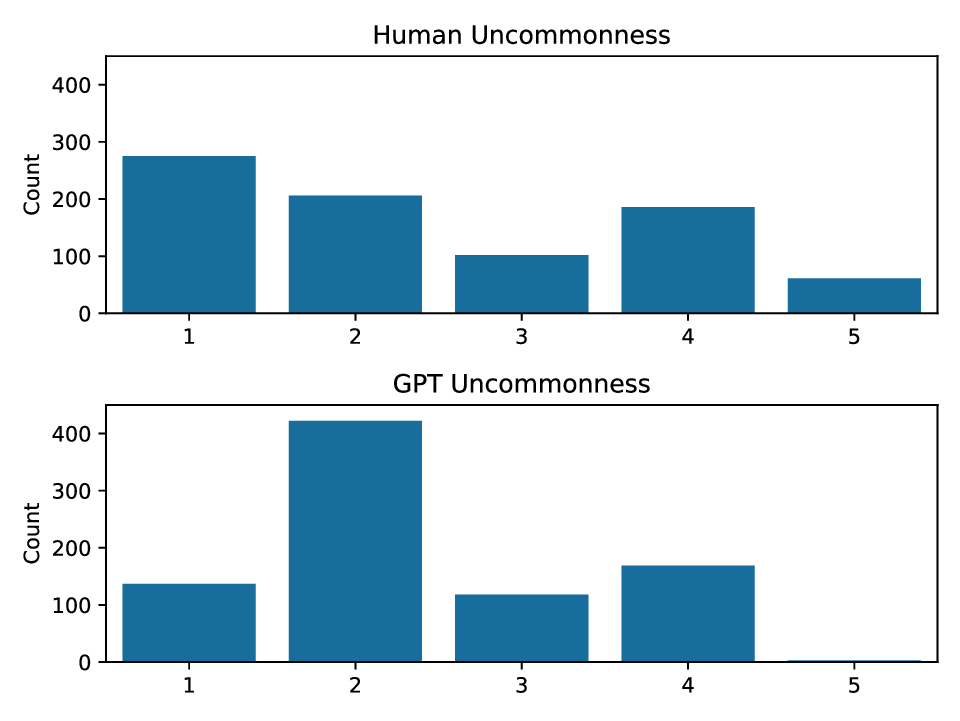}
    }
    \caption{Rating distributions for cleverness, remoteness, and uncommonness for humans and \textsc{gpt-4o-mini}.}
    \label{fig:human_gpt_cleverness_remote_uncommon}
\end{figure*}

Figure \ref{fig:experiment_2_claude_correlations} summarizes differences in facet correlations for \textsc{claude-3.5-haiku}. Trends are similar to that of \textsc{gpt-4o-mini}, though magnitudes in correlation are not as strong in many cases. Figures \ref{fig:gpt_agreement} and \ref{fig:claude_agreement} report Kappa agreement among the facets for \textsc{gpt-4o-mini} and \textsc{claude-3.5-haiku} respectively. Trends are largely similar to the Pearson correlations: agreement still rises sharply among all pairs in the example condition (pointing to homogenization) and cleverness is still the weakest predictor of originality for \textsc{gpt-4o-mini} (though the opposite is true for \textsc{claude-3.5-haiku}).  Rating distributions for cleverness, remoteness, and uncommonness for humans compared against \textsc{gpt-4o-mini} are reported in Figure \ref{fig:human_gpt_cleverness_remote_uncommon}.

\section{Additional Explanation Results}
\begin{figure*}[t]
    \centering
    \footnotesize
    \subfloat{
        \includegraphics[width=0.9\textwidth]{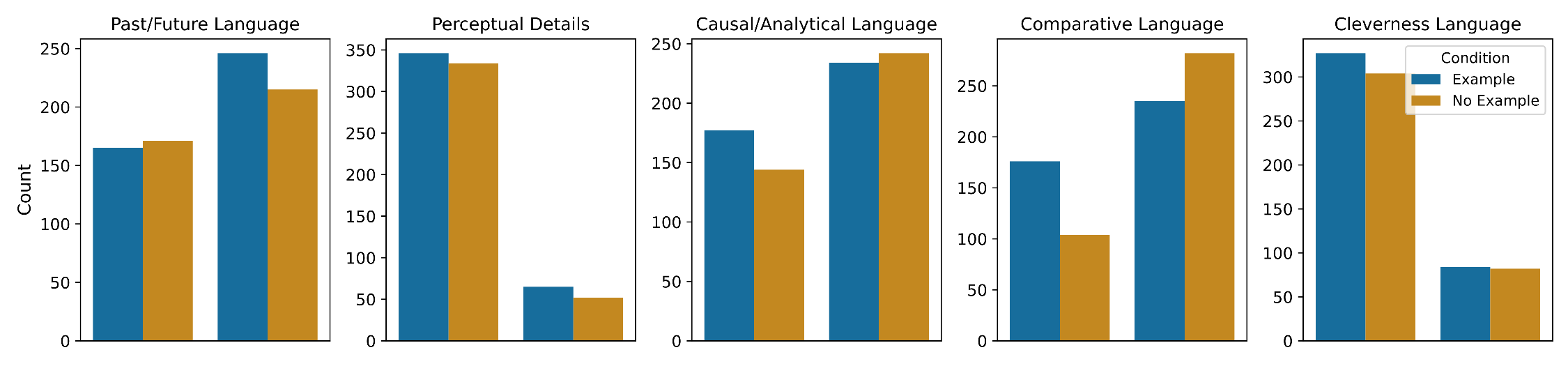}
    }
    \hfill
    \subfloat{
        \includegraphics[width=0.9\textwidth]{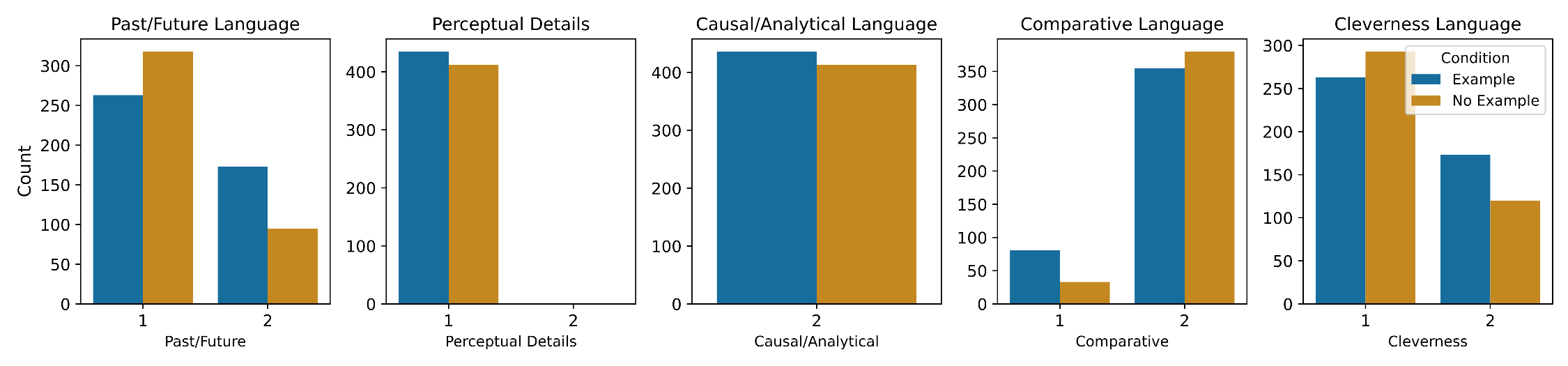}
    }
    \hfill
    \subfloat{
        \includegraphics[width=0.9\textwidth]{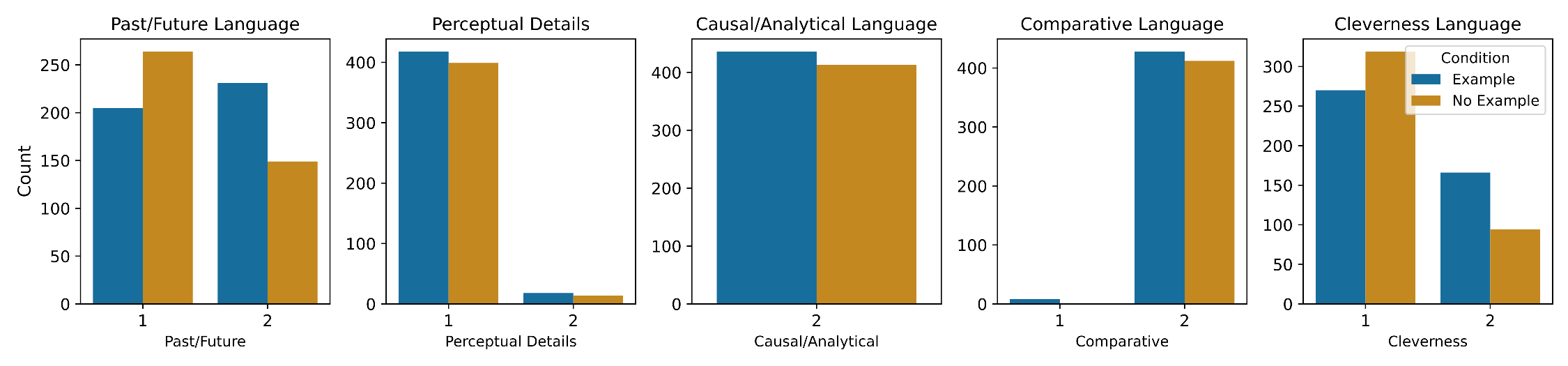}
    }
    \hfill
    \subfloat{
        \includegraphics[width=0.9\textwidth]{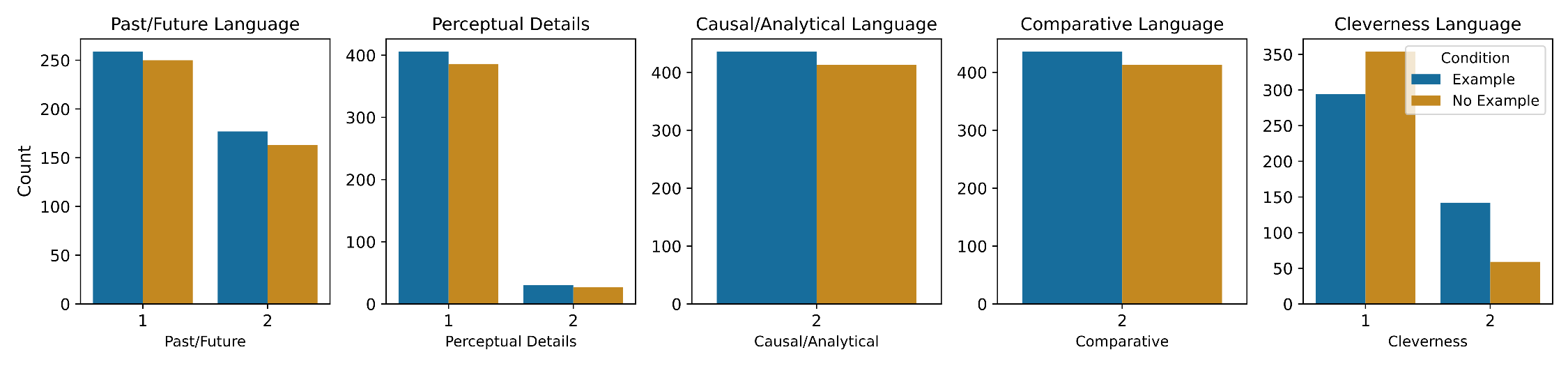}
    }
    \caption{Comparisons between linguistic marker use for all populations and LLM raters not reported in the main paper. Row 1: human explanations rated by \textsc{claude-3.5-sonnet}. Row 2: \textsc{gpt-4o-mini} explanations rated by \textsc{claude-3.5-sonnet}. Row 3: \textsc{claude-3.5-haiku} explanations rated by \textsc{claude-3.5-sonnet}. Row 4: \textsc{claude-3.5-haiku} explanations rated by \textsc{gpt-4o}.}
    \label{fig:all-liwc-analysis}
\end{figure*}

Results from \textsc{claude-3.5-haiku}, and from \textsc{claude-3.5-sonnet} for both human and \textsc{gpt-4o-mini} explanations, are presented in Figure \ref{fig:all-liwc-analysis}. For \textsc{claude-3.5-haiku}, both \textsc{claude-3.5-sonnet} and \textsc{gpt-4o} found a significant difference in past/future language use (claude U = 105253.5, p $<$ 0.001; GPT U = 105048.5, p $<$ 0.001), comparative language (claude U = 88600, p $<$ 0.05; GPT U = 80501.5, p $<$ 0.001), and cleverness (claude U = 103821, p $<$ 0.001; GPT U = 99598.5, p $<$ 0.01), but found no significant difference for perceptual details as a function of condition.

\section{Explanation Qualitative Analysis}
We include examples of human and \textsc{gpt-4o-mini} explanations for responses from the problem ``assist people with learning impairments retain information'' in Table \ref{tab:examples}. We find that human explanations tend to be shorter, more egocentric (e.g., ``I think''), and more variable in their structure and areas of focus. LLM explanations are more formulaic and make more direct references to prior ideas. We also find that LLMs sometimes address cleverness, remoteness, or uncommonness in their explanations, though in our prompt we emphasis that explanations should focus on originality, which supports the idea that LLMs homogenize these facets with originality.

\begin{table*}[htb]
\centering
\footnotesize
\caption{Sample human (first three) and \textsc{gpt-4o-mini} (last three) explanations for the same prompt.}
\begin{tabular}{c}
Explanations      \\ \hline                                                                                                                                                                                                                                                                                                                                                                          \\
\begin{tabular}[c]{@{}c@{}}I think this idea implements gamification and behavioral conditioning in an interesting way. \\ I'm not sure how adaptive it would be for learning impairments which can be varied. \end{tabular} \\ \hline    
\begin{tabular}[c]{@{}c@{}}This is original because it mimics the human encoding process for information. \\ This will also help many people with disabilities or speech impairments, \\ as they can listen to recordings as many times as needed. \end{tabular} \\ \hline
\begin{tabular}[c]{@{}c@{}}Said and done. We would need new and improved writing tools to assist, \\ but now that we have CGPT-4 and all these included advancements within said tech, \\ it's trying to find what else can work because right now there really isn't anything new and improved, \\ unless it's expensive and most people haven't heard of it. \end{tabular} \\ \hline

\\
\begin{tabular}[c]{@{}c@{}}This solution is original as it combines a common tool (flashcards) with a unique application (phone lock screen)\\ to create a novel way to reinforce learning. The integration of learning into a daily-use device is a clever approach, \\ though the concept of using a lock screen for educational purposes is somewhat familiar, which affects its remoteness. \end{tabular} \\ \hline
\begin{tabular}[c]{@{}c@{}}This solution is moderately original as it builds on existing educational technology concepts but \\ introduces a specific focus on time management and personalized learning. \\ While the idea of using apps for learning is common, \\ the clever integration of gauging time spent on topics adds a thoughtful layer to the approach.\end{tabular} \\ \hline
\begin{tabular}[c]{@{}c@{}}The idea of using art as a tool for learning is not particularly original, \\ as it has been explored in various educational contexts. While art can enhance memory retention, \\ the concept itself is common and lacks innovative engineering applications or methods. \end{tabular} \\ \hline
\end{tabular}
\label{tab:examples}
\end{table*}

\end{document}